%% file: tmi.tex
\documentclass[journal,twoside,web]{ieeecolor}
\usepackage{tmi}
\usepackage{cite}
\usepackage{amsmath,amssymb,amsfonts}
\usepackage{graphicx}
\usepackage{textcomp}
\usepackage{multirow}
\usepackage{tabularx}
\usepackage{booktabs}   
\usepackage{algorithm}
\usepackage[table,usenames,dvipsnames]{xcolor}
\usepackage{algpseudocode}

\def\BibTeX{{\rm B\kern-.05em{\sc i\kern-.025em b}\kern-.08em
    T\kern-.1667em\lower.7ex\hbox{E}\kern-.125emX}}
\begin{document}
\title{LLM-driven Medical Report Generation via Communication-efficient Heterogeneous Federated Learning}
\author{Haoxuan Che$^{*}$,\IEEEmembership{Student Member, IEEE}, Haibo Jin$^{*}$, Zhengrui Guo$^{*}$, Yi Lin, \IEEEmembership{Student Member, IEEE}, Cheng Jin,\IEEEmembership{Student Member, IEEE}, and Hao Chen$^{\dagger}$, \IEEEmembership{Senior Member, IEEE}
\thanks{This work was supported by the Pneumoconiosis Compensation Fund Board, HKSAR (Project No. PCFB22EG01), HKUST 30 for 30 Research Initiative Scheme (Project No. FS111), the Research Grants Council of the Hong Kong (Project Reference Number: T45-401/22-N) and the Project of Hetao Shenzhen-Hong Kong Science and Technology Innovation Cooperation Zone (HZQB-KCZYB-2020083).}
\thanks{H. Che, H. Jin, Z. Guo, Y. Lin, and C. Jin are with the Department of Computer Science and Engineering at the Hong Kong University of Science and Technology University, Hong Kong SAR, China. (e-mail: \{hche, hjinag, zguobc, yi.lin, cheng.jin\}@connect.ust.hk).}
\thanks{H. Chen is with the Department of Computer Science and Engineering, Department of Chemical and Biological Engineering and Division of Life Science, Hong Kong University of Science and Technology, Hong Kong, China (e-mail: jhc@cse.ust.hk).}
\thanks{$^{*}$ denotes equal contribution to this work.}
\thanks{$^{\dagger}$ indicates corresponding author.}
}

\maketitle

\input{0_abstract}
\input{1_introduction}
\input{2_relatedworks}
\input{3_method}
\input{4_exp}
\input{5_conclusion}

\bibliographystyle{IEEEtran}
\bibliography{ref}
\end{document}

%% file: 0_abstract.tex
\begin{abstract}
Large Language Models (LLMs) have demonstrated significant potential in Medical Report Generation (MRG), yet their development requires large amounts of medical image-report pairs, which are commonly scattered across multiple centers. Centralizing these data is exceptionally challenging due to privacy regulations, thereby impeding model development and broader adoption of LLM-driven MRG models. 
To address this challenge, we present FedMRG, the first framework that leverages Federated Learning (FL) to enable privacy-preserving, multi-center development of LLM-driven MRG models, specifically designed to overcome the critical challenge of communication-efficient LLM training under multi-modal data heterogeneity.
To start with, our framework tackles the fundamental challenge of communication overhead in federated LLM tuning by employing low-rank factorization to efficiently decompose parameter updates, significantly reducing gradient transmission costs and making LLM-driven MRG feasible in bandwidth-constrained FL settings. Furthermore, we observed the dual heterogeneity in MRG under the FL scenario: varying image characteristics across medical centers, as well as diverse reporting styles and terminology preferences. 
To address the data heterogeneity, we further enhance FedMRG with (1) client-aware contrastive learning in the MRG encoder, coupled with diagnosis-driven prompts, which capture both globally generalizable and locally distinctive features while maintaining diagnostic accuracy; and (2) a dual-adapter mutual boosting mechanism in the MRG decoder that harmonizes generic and specialized adapters to address variations in reporting styles and terminology. 
Through extensive evaluation of our established FL-MRG benchmark, we demonstrate the generalizability and adaptability of FedMRG, underscoring its potential in harnessing multi-center data and generating clinically accurate reports while maintaining communication efficiency.
\end{abstract}

%% file: 1_introduction.tex
\begin{figure}
    \centering
    \includegraphics[width=\linewidth]{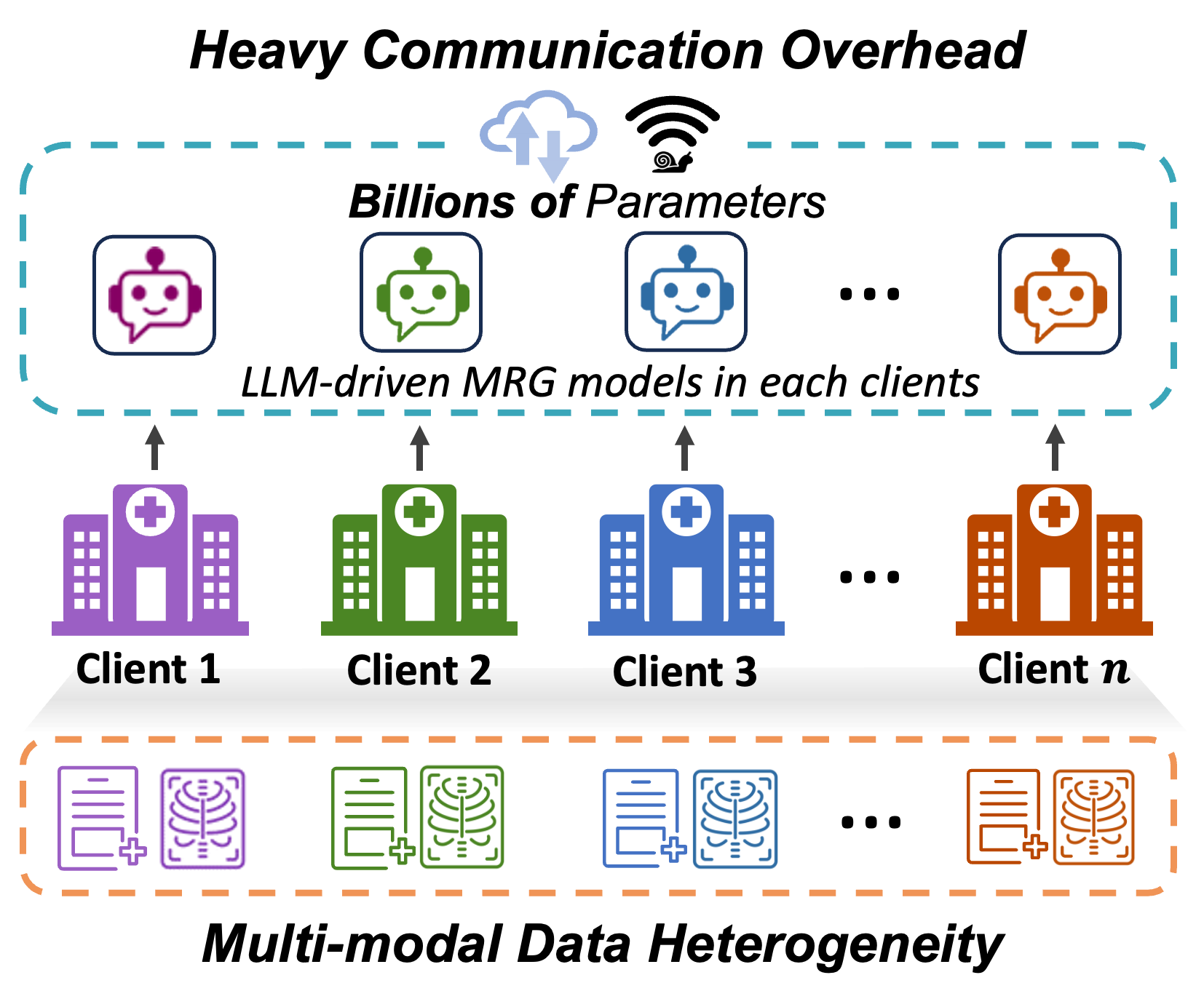}
    \vspace{-8mm}
    \caption{Training a LLM-driven MRG model in FL settings face challenges: 1) heavy communication overheads, and 2) multi-modal data heterogeneity among centers.}
    \vspace{-3mm}
    \label{fig:hierachy-domain-shifts}
\end{figure}

\section{Introduction}
\label{sec:introduction}
Recently, Large Language Models (LLMs) have emerged as a promising approach for Medical Report Generation (MRG), enabling automatic creation of detailed clinical descriptions from medical images \cite{wang2023r2gengpt,lee2023llm,yang2023medxchat,Omkar2023XrayGPT,chen2024dia,liu2024bootstrapping,jin2023promptmrg,che2023dgdr,che2022learning,che2023iqad}. 
With their extensive pre-trained knowledge and advanced reasoning capabilities \cite{touvron2023llama}, LLMs demonstrate superior performance in capturing subtle clinical findings and generating linguistically coherent reports, offering significant value in reducing radiologists' workloads and improving diagnostic efficiency.
However, the advancement of LLM-driven MRG faces substantial challenges, primarily due to the fragmented and privacy-restricted nature of medical report data across healthcare institutions \cite{jing2018automatic,sun2021pain,zhang2024unpaired,dai2024deep,che2025feddag}.

Despite the growing volume of medical data, its utility is often constrained by stringent privacy regulations and disparate legal frameworks across regions and nations \cite{price2019privacy}. 
Sharing medical imaging and report data between centers remains legally prohibited in many cases, and even when sharing is permissible, the transmission and storage of substantial datasets pose significant technical challenges \cite{johnson2019mimic}. Consequently, individual medical centers face an acute scarcity of training data when attempting to develop LLM-driven MRG models independently.
This situation is further complicated by the massive parameter count of modern LLMs—typically reaching billions—which inherently requires larger and more diverse training datasets to achieve optimal performance \cite{touvron2023llama}.
This creates a paradoxical situation: while advanced LLMs offer superior capabilities for medical report generation, they simultaneously demand data volumes that exceed what single institutions can typically provide.
Beyond training challenges, these data limitations directly impact clinical deployment safety \cite{sheller2020federated}. 
Limited exposure to diverse medical cases can increase the generation hallucination of LLMs (a phenomenon where LLMs produce text that is factually incorrect or unfounded despite appearing fluent and coherent), which poses significant risks when such models are applied in clinical settings \cite{clusmann2023future}.
These interconnected challenges underscore the urgent need for innovative approaches that can leverage distributed medical data while preserving privacy and addressing the unique requirements of LLM-driven MRG systems.

Incorporating federated learning (FL) presents a promising pathway for addressing the data challenges inherent in LLM-driven MRG \cite{rieke2020future}.
As a distributed learning paradigm that enables collaborative model training across multiple centers without direct data sharing \cite{mcmahan2017fedavg}, FL offers a potential solution for increasing data availability while preserving privacy in the sensitive medical domain.
However, deploying LLM-driven MRG within an FL framework introduces substantial technical challenges, particularly related to \textit{communication-efficient LLM training under multi-modal data heterogeneity}, as illustrated in Fig. \ref{fig:hierachy-domain-shifts}. One major obstacle is the massive communication overhead imposed by LLMs' billions of parameters.
For example, considering that 1000 clients collaborate for training a PaLM \cite{chowdhery2023palm} model and each client has a network with a speed of 1GB/second, \emph{only} the communication between a single client and server needs 552 hours, and the server network speed \emph{must be} 1000GB/second for serving all clients.
Such prohibitive communication requirements render conventional federated approaches infeasible for LLM-driven MRG without specialized efficiency techniques.

On the other hand, the MRG task presents unique challenges for federated LLM training due to heterogeneity that simultaneously spans both image and report data \cite{chen2023feddat}. Unlike single-modality FL problems, this dual heterogeneity creates a compound effect where variations in one modality can amplify inconsistencies in the other, undermining conventional FL approaches \cite{munro2020multi}.
The essence of this challenge lies in the diverse and sometimes conflicting nature of parameter distributions within the model's trainable components, such as the image encoder and text decoder, across different client sites.
Medical centers naturally exhibit significant variations in imaging equipment, acquisition protocols, reporting templates, and terminology preferences—creating a challenging environment where naive federation can lead to conflicting parameter updates.
When aggregated at the server, these conflicting updates risk model degradation, instability, or even complete collapse \cite{heim2023towards}.

In this paper, we introduce Federated Learning for Medical Report Generation, dubbed FedMRG, a novel framework designed to communication-efficiently adapt LLMs for the MRG task amidst multi-modal data heterogeneity across medical centers. Our approach enables cooperative development of LLM-driven MRG models while addressing the critical challenges of data privacy and cross-institutional variability. Specifically, FedMRG first incorporates low-rank factorization to decrease the trainable parameter size of LLMs for each client, achieving significant communication cost reduction in the FL scenario. Further, we propose two integral components to mitigate the challenge of FL data heterogeneity: hierarchical contrasting and prompting (HCP) for the image encoder and dual-adapter mutual boosting (DMB) for the text decoder.

\textit{On the image encoder}, HCP employs a two-tier approach that combines self-supervised contrastive learning in local clients with negative samples from a global memory bank, and further tokenizes diagnosis predictions to guide report generation, ensuring both clinical precision and relevance.
\textit{On the text decoder}, DMB integrates two complementary LLM adapters: one optimized for generic global knowledge and another specialized for local data nuances, facilitating synergistic improvement through mutual knowledge transfer. To rigorously evaluate our approach, we developed FL-MRG, a comprehensive benchmark that simulates real-world client data heterogeneity. Extensive experiments demonstrate FedMRG's superior generalization and effectiveness, highlighting its potential to benefit the development of real-world MRG models.
Our contributions can be highlighted in three folds:

\begin{enumerate}
    \item We present the first framework that integrates FL with LLM-driven MRG, pioneering a privacy-preserving approach for multi-center medical report generation that specifically addresses the dual challenge of communication efficiency and multi-modal data heterogeneity.
    \item We propose FedMRG, a novel framework with several delicately designed modules: it first tackles communication overhead through parameter-efficient low-rank factorization, then addresses multi-modal data heterogeneity via Hierarchical Contrasting and Prompting (HCP) for image encoding and Dual-adapter Mutual Boosting (DMB) for text decoding.
    \item We establish FL-MRG, the first comprehensive benchmark for federated medical report generation that simulates realistic cross-center heterogeneity. Through extensive experiments comparing 14 state-of-the-art methods and detailed ablation studies, we demonstrate FedMRG's superior performance in both communication efficiency and clinical accuracy.
\end{enumerate}

%% file: 2_relatedworks.tex
\section{Related Works}
\label{sec:rel}
\subsection{Medical Report Generation}
Medical Report Generation (MRG) aims to automate the creation of narrative text from medical images \cite{monshi2020deep}. It presents unique challenges beyond standard image captioning, particularly in identifying clinical abnormalities and generating lengthier reports \cite{jin2023promptmrg}.
Recognizing its critical role in healthcare, researchers have developed advanced approaches spanning memory modules \cite{chen2020generating,yang2023radiology,guo2024histgen}, knowledge graphs \cite{zhang2020radiology,liu2021exploring,li2023dynamic,huang2023kiut}, multi-task learning frameworks \cite{jing2018automatic,wang2022automated,wang2021self,yan2021weakly,yan2022clinical,tanida2023interactive}, and mix-of-expert architectures \cite{wang2023metransformer}.
Recent trends show significant interest in adapting LLMs for MRG tasks \cite{wang2023r2gengpt,lee2023llm,yang2023medxchat,Omkar2023XrayGPT,chen2024dia,liu2024bootstrapping,jin2023promptmrg,jin2025chain}, leveraging their comprehensive knowledge foundations.
Jin et al. \cite{jin2023promptmrg}, for instance, enhanced diagnostic accuracy through diagnosis-driven prompts guiding the decoding process.
Building upon miniGPT-4 \cite{zhu2023minigpt}, pioneering contributions from Liu et al. \cite{liu2024bootstrapping} and Wang et al. \cite{wang2023r2gengpt} demonstrated the adaptation of pre-trained LLMs to MRG tasks.
Wang et al. \cite{wang2023r2gengpt} specifically investigated three fine-tuning strategies for LLMs, establishing their effectiveness for MRG applications.
Concurrently, Liu et al. \cite{liu2024bootstrapping} developed a bootstrapping approach for LLMs in MRG using in-domain instance induction and coarse-to-fine decoding, with similar concepts explored in \cite{chen2024large}.

Despite these advancements, a fundamental challenge persists: limited availability of medical data directly constrains the precision of generated reports, particularly for data-hungry LLM-driven MRG systems.
This limitation underscores the relevance of integrating FL into LLM-driven MRG, where we achieve it by introducing FedMRG.
Unlike previous MRG research, our work represents the first exploration of incorporating FL into MRG, offering transformative potential for addressing persistent data limitations.
By applying the FL paradigm to LLM-driven MRG, we harness decentralized computing to tackle data challenges effectively.
FedMRG enables collaborative model training across multiple centers while maintaining data localization, thereby expanding available datasets for LLM-driven MRG models without requiring data centralization, while simultaneously addressing data privacy and security concerns.

\subsection{Federated Large Model Adaptation}
Federated Learning (FL) emerges as a privacy-conscious learning paradigm, where clients, following strict data privacy standards, collaborate on model training without the necessity of data sharing.
Owing to its nature of boosting collaboration and providing more data to foundation models, recently, a lot of pioneered works have focused on introducing FL into foundation model construction.
They typically focus on designing communication strategies and introducing parameter-efficient fine-tuning (PEFT) methods to reduce the huge communication overhead raised by the foundation model parameter size \cite{wang2023cocktailsgd,guo2023promptfl,yang2023efficient,li2024global,su2024federated,chen2023feddat,zhang2024towards,gu2024mix,sun2024improving}.
For example, CocktailSGD \cite{wang2023cocktailsgd} combines different classic communication strategies, including random sparsification, top-K sparsification, and quantization, with the local SGD algorithm to achieve superior communication efficiency.
As for leveraging parameter-efficient fine-tuning methods, previous works mainly discussed the utilization of methods such as the prompt learning \cite{guo2023promptfl,yang2023efficient,li2024global,su2024federated}, the MLP-based adapter \cite{chen2023feddat} and the low-rank adaptation \cite{zhang2024towards,gu2024mix,sun2024improving} to reduce the number of trainable parameters. For example, the low-rank adaptation (LoRA) technique \cite{hu2021lora} could be simply integrated into the federated learning setting to achieve communication-efficient training. Similar adaptations from general domain to federated learning includes Fed-Prefix \cite{li2021prefix}, Fed-Prompt \cite{lester202prompt}, Fed-AdaLoRA \cite{zhang2023adlora}, and Fed-Vera \cite{kopiczko2023vera}. To further unlock the low-rank constraints under the federated learning setting, FedPara \cite{hyeon2021fedpara} re-parameterizes weight parameters of layers using low-rank weights followed by the Hadamard product. Although these methods can achieve adorable communication efficiency and performance improvements, they are sensitive to the intrinsic data heterogeneity in FL.

In the MRG task, the situation is more challenging since there are two data modalities at the same time, which requires re-designing another advanced framework instead of simply applying existing PEFT methods in federated scenarios. Different from previous works \cite{hu2021lora,li2021prefix,lester202prompt,zhang2023adlora,kopiczko2023vera,hyeon2021fedpara}, our paper emphasizes the importance of addressing data heterogeneity when adapting LLMs for the MRG task while maintaining communication efficiency. While this challenge has been partially explored in FedDAT \cite{chen2023feddat}, which also considers data heterogeneity mitigation and implements a dual adapter design, our framework is specifically designed for medical report generation with distinct technical contributions, \textit{i.e.}, our approach integrates specialized encoder-decoder components that handle both visual feature heterogeneity through client-aware contrastive learning and text heterogeneity via a dual-adapter mechanism tailored for medical terminology and reporting styles across different medical centers.


%% file: 3_method.tex
\begin{figure*}
    \centering
    \includegraphics[width=\linewidth]{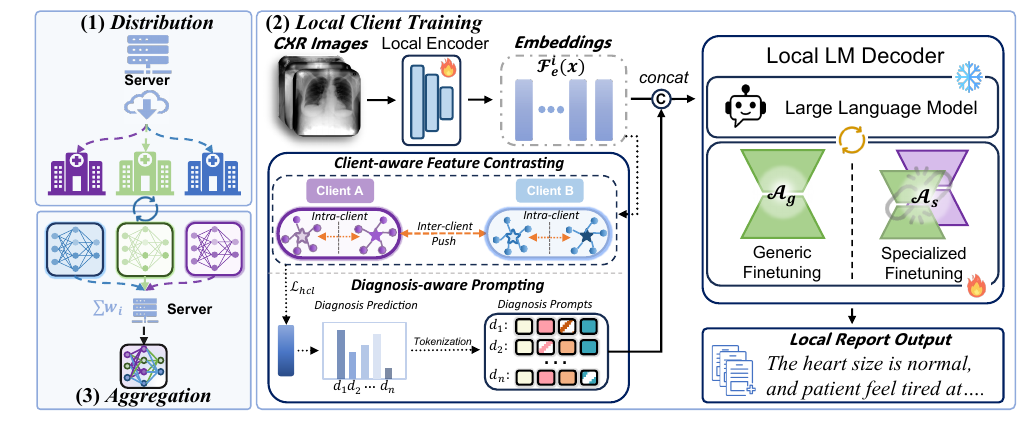}
    \vspace{-6mm}
    \caption{The framework of FedMRG with three stages after model initialization: (1) \textbf{Distribution}: The server distributes initialized models to clients. (2) \textbf{Local Client Training}: Clients train models using our two key modules: \textbf{Hierarchical Contrasting and Prompting (HCP)} addresses image heterogeneity through client-aware feature contrasting (enabling inter- and intra-client knowledge sharing) and diagnosis-aware prompting (converting disease predictions into tokens [BLA]/[POS]/[NEG]/[UNC] to guide generation); \textbf{Dual-adapter Mutual Boosting (DMB)} tackles text heterogeneity using a generic adapter $\mathcal{A}_g$ for global patterns and a specialized adapter $\mathcal{A}_s$ for client-specific reporting styles. (3) \textbf{Aggregation}: Only encoder parameters and generic adapters are uploaded to reduce communication overhead, which the server aggregates and redistributes. This framework effectively addresses both communication efficiency and multi-modal data heterogeneity challenges in federated MRG.}
    \vspace{-3mm}
    \label{fig:framework}
\end{figure*}

\section{Methodology}
\label{sec:method}
\subsection{Communication-efficient Federated MRG System}
Training an LLM-driven MRG model in federated settings presents a formidable challenge regarding communication efficiency.
The massive parameter count of LLMs renders training with conventional federated algorithms, such as FedAVG \cite{mcmahan2017fedavg}, prohibitively expensive in terms of communication overhead.
This communication bottleneck effectively makes the deployment of LLM-driven MRG models impractical within federated environments.
To address this critical constraint, we introduce FedMRG, a novel communication-efficient federated MRG framework. FedMRG strategically leverages LLMs to enhance MRG capabilities while simultaneously minimizing both computational and communication resource demands.

\subsubsection{Learning Objectives}
In our framework, we consider a federation of $n$ client domains, denoted as $\mathcal{D} = \{D_1, D_2, \ldots, D_n\}$. Each client domain $D_i$ maintains a local dataset of paired medical images and reports ${(x_k^i, r_k^i)}_{k=1}^{N^i}$, sampled from its domain-specific distribution $(\mathcal{X}^i, \mathcal{R}^i)$. Our primary objective is to develop a global MRG model that effectively harnesses the collective knowledge from all distributed clients. Through this federated approach, we aim to construct a model that achieves robust generalization across both source domains ($\mathcal{D}$) and unseen domains ($\mathcal{U}$), while minimizing the communication overhead inherent in distributed learning systems.

\subsubsection{Base Architecture}
In each client, we establish a vanilla MRG encoder-decoder architecture as our foundation (Fig. \ref{fig:framework}), comprising a visual encoder $\mathcal{F}_{e}$ that extracts visual features from an image $x$ and a frozen LLM decoder $\mathcal{F}_{d}$ with trainable components that generates reports $r$ conditioned on these visual features.
We denote visual feature extraction as $\mathcal{F}_{e}(x) = \{f_1, f_2, \ldots, f_S\}$, where $f_i \in \mathbb{R}^C$ represents a feature patch, $C$ denotes the feature dimension, and $S$ is the number of patches.
We define a report with length $T$ as $r = \{v_1, v_2, \ldots, v_T \}$, $v_i \in \mathbb{V}$, where $\mathbb{V}$ is the vocabulary.
The decoding process can be formulated as $v_t = \mathcal{F}_{d} (f_1, \ldots, f_S, v_1, \ldots, v_{t-1})$, where $v_t$ is the token predicted at time step $t$.
Given the prediction $r^\prime$ and ground truth report $r$, the loss function for report generation employs language modeling loss $\mathcal{L}_{LM}$ as
\begin{equation}
    \mathcal{L}_{LM} (r^\prime, r) = - \sum_{t=1}^T \log p(v_t| v_1, v_2, \ldots, v_{t-1}). 
\end{equation}


\subsubsection{Enabling Communication Efficiency through Low-rank Adaptation}
\label{sec:lora}
Inspired by \cite{hu2022lora}, we incorporate low-rank adaptation (LoRA) into our framework. LoRA introduces lightweight adapters that efficiently modify the pre-trained LLM weights without requiring full model updates. For each module $m$, LoRA defines an adapter $\Delta \theta_m = \Delta \theta_m^b \Delta \theta_m^a$, where $\Delta \theta_m^b$ and $\Delta \theta_m^a$ are low-rank matrices. These adapters transform the original weight matrix $W_m$ as:
\begin{equation}
W_m' = W_m + \Delta \theta_m^b \Delta \theta_m^a
\end{equation}
This approach delivers substantial advantages for federated MRG by dramatically reducing parameter updates and minimizing communication overhead. The low-rank structure of the adapters enables efficient adaptation of the LLM to medical domain specifics, while ensuring resource-efficient deployment across distributed healthcare institutions. The compressed parameter space not only reduces communication costs but also mitigates overfitting risks, making it particularly suitable for federated learning environments where data privacy and communication efficiency are paramount concerns.

In subsequent sections, we elaborate on our specific designs for the image encoder (Sec. \ref{sec:hclr}), text decoder (Sec. \ref{sec:HEFB}), and present our comprehensive framework (Sec. \ref{sec:overall}).

\subsection{Hierarchical Contrasting and Prompting}
\label{sec:hclr}
While our LoRA-based LLM tuning substantially reduces communication costs, making LLM training feasible in federated settings, the inherent isolation of client image data introduces another significant challenge: \textit{learning globally generalizable features while preserving client-specific characteristics, particularly when processing heterogeneous medical image data across centers.} This challenge is especially pronounced in medical report generation, where diseases, despite sharing common classifications and symptoms, exhibit considerable visual variations across different patients and medical centers \cite{johnson2019mimic}. For example, while pneumonia may have standardized diagnostic criteria, its visual presentation in chest X-rays can vary dramatically due to patient-specific factors and center-specific imaging protocols. 

This duality — requiring both the capture of center-specific lesion variations and maintenance of consistent diagnostic interpretations — motivates our Hierarchical Contrasting and Prompting (HCP) approach, drawing inspiration from \cite{jin2023promptmrg,khosla2020supervised}. HCP addresses this challenge through two complementary mechanisms: client-aware feature contrasting to learn discriminative visual representations that encapsulate both local variations and global patterns, and diagnosis-aware prompting to provide structured guidance for the medical report generation process, ensuring both clinical accuracy and contextual relevance. These modules operate synergistically to overcome image heterogeneity across distributed client sites.

\subsubsection{Client-aware Feature Contrasting}
At the feature level, HCP aims to incorporate the unique characteristics associated with individual samples through contrastive learning. 
Beyond the self-supervised contrastive learning on local clients, we introduce a global memory bank $\mathcal{M}$ to achieve client-aware feature representation learning by providing negative samples derived from other clients.
$\mathcal{M}$ stores features randomly sampled from each local client and updates and shares these features through parameter communication.
Specifically, within a mini-batch, let $i \in I \equiv \{1...N\}$ be the index of arbitrary local samples, $j(i) \in J \equiv \{1^\prime...N^\prime\}$ be the index of samples originating from the same sample $i$, and $k \in M \equiv \{\hat{1}...\hat{N}\}$ be the index of samples whose features are sampled from $\mathcal{M}$. 
We denote $f^\cdot_{avg}$ as the $l_2$-normalized pooling features.
The contrastive loss $\mathcal{L}_{hcl}$ for sample $i$ is then denoted as:
\begin{equation}
    \mathcal{L}_{hcl}^i = - \log \frac{\exp(f_{avg}^i \cdot f^{j(i)}_{avg} / \tau)}{\sum_{a\in A(i)} \exp(f^i_{avg} \cdot f^a_{avg} / \tau)},
\end{equation}
where the $\cdot$ symbol denotes the inner product, $\tau$ is the temperature parameter, and $A(i) \equiv (J \cup I \cup M) / \{i, i^\prime\}$.

\subsubsection{Diagnosis-aware Prompting}
To capture the similarity among reports under the same diagnosis, we convert the diagnosis prediction as the input prompt for the text decoder, providing robust clinical guidance for text generation. Specifically, our framework includes a disease classification branch that performs 4-class classification (Blank, Positive, Negative, Uncertain) for each disease. We leverage the cross-entropy loss $\mathcal{L}_{\text{CE}}$ to train this branch on disease labels. The classification labels can be obtained with CheXbert \cite{irvin2019chexpert} by converting reports into 14 predefined disease labels, and the training can be done with standard cross-entropy loss $\mathcal{L}_{\text{CE}}$.

During training, the ground-truth disease labels are converted into four distinct token prompts: [BLA], [POS], [NEG], and [UNC], which are added to the decoder's vocabulary. These diagnosis-based prompts are utilized as part of the input in the text decoder. The text decoder then attends to both the visual features and these diagnosis-specific prompts during report generation, allowing it to incorporate diagnostic information when generating clinically accurate reports explicitly. During inference, the diagnostic predictions from the classification branch are converted into these prompt tokens automatically, helping the model generate reports that are not only linguistically coherent but also clinically accurate. This approach is particularly effective in the federated learning setting, where maintaining diagnostic consistency across heterogeneous client data is challenging.

In essence, HCP concentrates on optimizing the MRG encoder architecture, specifying the precise types of features and prompts required to overcome the unique image heterogeneity challenges presented by the FL setting.
Features capture fine-grained visual information, while prompts encapsulate coarse-grained diagnostic context. Our approach is methodically designed to be comprehensive, addressing both levels of information granularity effectively. By seamlessly integrating these complementary aspects, HCP enhances the model's capacity to extract distinctive yet generalizable features while ensuring consistent performance across diverse client datasets.

\subsection{Dual-adapter Mutual Boosting}
\label{sec:HEFB}
While low-rank adapters (see Sec. \ref{sec:lora}) substantially reduce communication costs during LLM tuning, they exhibit limitations in simultaneously capturing global patterns and local variations across institutions, \textit{i.e.,} text heterogeneity.
As previously established, healthcare facilities serve diverse patient populations and maintain distinct documentation protocols and clinical specializations, resulting in highly variable report structures and terminology.
To address this challenge, we propose the Dual-adapter Mutual Boosting (DMB) module, which implements a generic adapter for encoding global reporting patterns while minimizing communication overhead, alongside a specialized adapter for preserving client-specific reporting styles without necessitating parameter sharing. DMB facilitates bidirectional knowledge transfer between these adapters through a mutual boosting mechanism, enabling global insights to inform local adaptations and vice versa. This architecture effectively balances communication efficiency with the capacity to accommodate heterogeneous reporting practices across distributed medical centers and institutions.

\subsubsection{Dual-adapter Design}
\begin{figure}
    \centering
    \includegraphics[width=1\linewidth]{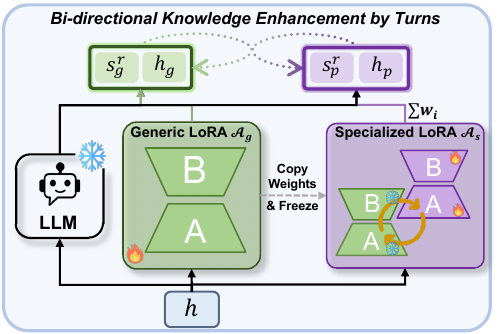}
    \caption{The DMB involves two adapter configurations and tuning strategies, and bi-directional knowledge enhancement via distillation.}
    \label{fig:HEFB}
    \vspace{-3mm}
\end{figure}

As illustrated in Fig. \ref{fig:HEFB}, DMB integrates two distinct adapters into each layer of $\mathcal{F}_d$: a generic adapter $\mathcal{A}_g$ and a specialized expert adapter $\mathcal{A}_s$.
$\mathcal{A}_g$ is designed to encode global knowledge patterns and participates in cross-client communication.
$\mathcal{A}_s$ concentrates on the representation of client-specific knowledge and remains localized without participation in the aggregation of models. The fundamental distinction between these adapters resides in their architecture and operational functions: $\mathcal{A}_g$ is implemented as a standard LoRA module and optimized through generic fine-tuning, aggregating global knowledge via conventional federated averaging algorithms.
$\mathcal{A}_s$ comprises two complementary LoRA modules, trained through specialized fine-tuning. 
The first module functions as a reservoir for global knowledge by inheriting weights from $\mathcal{A}_g$ at the commencement of each local training round before being frozen.
The second module extends this global foundation to capture local client-specific nuances, facilitating specialized knowledge acquisition.
This dual-adapter configuration enables FedMRG to maintain an optimal balance between global insights and domain-specific characteristics.

\subsubsection{Mutual Boosting Mechanism}
Within our training framework, we implement a mutual boosting mechanism to facilitate bidirectional knowledge enhancement between $\mathcal{A}_g$ and $\mathcal{A}_s$. This mechanism ensures that local insights systematically inform the global perspective and vice versa, establishing an effective conduit for knowledge exchange.
Inspired by previous works \cite{deng2023unlocking, chen2022bridging}, which demonstrate that specialized models can substantially contribute to generalizable generic models, we implement this mechanism through knowledge distillation.
We denote the report generation logits as $s^r_s$ and $s^r_g$, and the final layer hidden states as $h_s$ and $h_g$ from the models incorporating $\mathcal{A}_s$ and $\mathcal{A}_g$, respectively.
Following \cite{sanh2019distilbert}, we use cosine embedding loss $\mathcal{L}_{cos}$ and KL divergence loss $\mathcal{L}_{KL}(\cdot||\cdot)$ for knowledge distillation:
  \begin{align}
      \mathcal{L}_{l2g} &= \mathcal{L}_{cos}(h_s, h_g) + \mathcal{L}_{KL}(s^r_g || s^r_s), \\
      \mathcal{L}_{g2l} &= \mathcal{L}_{cos}(h_g, h_s) + \mathcal{L}_{KL}(s^r_s || s^r_g),
  \end{align}
where $\mathcal{L}_{l2g}$ directs the transfer of local knowledge to the generic adapter, and $\mathcal{L}_{g2l}$ guides the transfer of global insights to the specialized adapter. During local training iterations, $\mathcal{A}_g$ and $\mathcal{A}_s$ are alternately activated and optimized using the language modeling loss $\mathcal{L}_{LM}$ along with the corresponding knowledge distillation losses.

Fundamentally, our DMB approach harnesses the complementary strengths of both global and local knowledge within a communication-efficient framework. By integrating specialized adapters and facilitating reciprocal knowledge transfer, we ensure the model maintains generalizability across shared patterns while preserving adaptability to client-specific nuances. This mutual boosting mechanism constitutes the cornerstone of our approach, enabling both parameter-efficient LLM adaptation and robust performance in the presence of text data heterogeneity.

\begin{algorithm}[t]
\caption{Pseudocode of FedMRG} \label{algo:llm}
\begin{algorithmic}[1]
    \State \textbf{Server Initialization:}
    \For{each client $i = 1, 2, \ldots, n$}
        \State Initialize a MRG model with $\mathcal{F}_e$ and $\mathcal{F}_d$ 
        \State Freeze parameters of $\mathcal{F}_d$
        \State Insert $\mathcal{A}_s$ and $\mathcal{A}_g$ into $\mathcal{F}_d$
        \State Distribute initialized model to client $i$
    \EndFor
    \Repeat
        \State \textbf{Local Client Training:}
        \For{each client $i = 1, 2, \ldots, n$ \textbf{in parallel}}
            \For{each mini-batch $j = 1, 2, \ldots, N^i$}
            \State Compute $\mathcal{L}_1 := \alpha \mathcal{L}_{HCL} + \mathcal{L}_{CE} + \beta \mathcal{L}_{l2g} + \mathcal{L}_{LM}$
            \State Optimize $\mathcal{F}_e$ and each $\mathcal{A}_g$ via $\mathcal{L}_1$
            \State Compute $\mathcal{L}_2 := \alpha \mathcal{L}_{HCL} + \mathcal{L}_{CE} + \beta \mathcal{L}_{g2l} + \mathcal{L}_{LM}$
            \State Optimize $\mathcal{F}_e$ and each $\mathcal{A}_s$ via $\mathcal{L}_2$
            \State Upload $\mathcal{F}_e$ and each $\mathcal{A}_g$ and $\mathcal{A}_s$ 
            \EndFor
        \EndFor
        \State \textbf{Server Aggregation and Distribution:}
        \State Aggregate clients updated $\mathcal{F}_e$ and $\mathcal{A}_g$ 
        \State Distribute the aggregated components
    \Until{maximum communication rounds reached}
\end{algorithmic}
\end{algorithm}

\subsubsection{Inference Strategy with Dual Adapters}
During the inference phase, our dual-adapter design provides flexibility based on the deployment scenario. For personalized inference within participating clients, both the generic adapter $\mathcal{A}_g$ and the client's specialized adapter $\mathcal{A}_s$ are activated and integrated through a specific mechanism. After the input passes through the frozen LLM backbone, the outputs from both adapters are combined using a weighted sum:

\begin{equation}
    \text{output} = \alpha \cdot \mathcal{A}_g(x) + (1-\alpha)\cdot\mathcal{A}_s(x),
\end{equation}

where $x$ is the hidden representation from the LLM backbone and $\alpha$ is a weighting parameter (set to 0.5 in our implementation) that balances global and client-specific knowledge. This integration mechanism ensures that the generated reports maintain consistency with global standards while preserving institution-specific terminology and reporting conventions.

For out-of-domain generalization scenarios, where the model is deployed at new medical centers without specialized adapters, only the generic adapter $\mathcal{A}_g$ is utilized. In this configuration, all adaptation is handled exclusively by the generic adapter, leveraging the collective knowledge aggregated across all training clients.

\subsection{Overall Framework of FedMRG}
\label{sec:overall}
{\color{black}
The pseudocode of the FedMRG algorithm is presented in Algo. \ref{algo:llm}. Fundamentally, FedMRG operates through four critical phases within a server-client architecture based on FedAvg \cite{mcmahan2017fedavg}. In the \textbf{initialization} phase, the server creates and distributes $n$ local MRG models to participating clients, with each model's decoder equipped with both generic and specialized adapters. During the \textbf{client training} phase, each client iteratively optimizes their model through: (1) adapter optimization using $\mathcal{L}_{LM}$ for basic report generation capabilities and $\mathcal{L}_{l2g}$, $\mathcal{L}_{g2l}$ for bidirectional knowledge transfer between adapters, and (2) encoder training utilizing hierarchical contrasting and diagnosis-aware prompting. Upon completion of local training, clients upload only their encoder parameters and generic adapter parameters to minimize communication overhead. In the \textbf{aggregation} phase, the server conducts parameter averaging on the received model components, followed by the \textbf{distribution} phase, where these aggregated parameters are transmitted back to reinitialize client-side models.}

%% file: 4_exp.tex
\section{Experiments}
\label{sec:experiment}
{\color{black}
\begin{figure}
    \centering
    \includegraphics[width=1\linewidth]{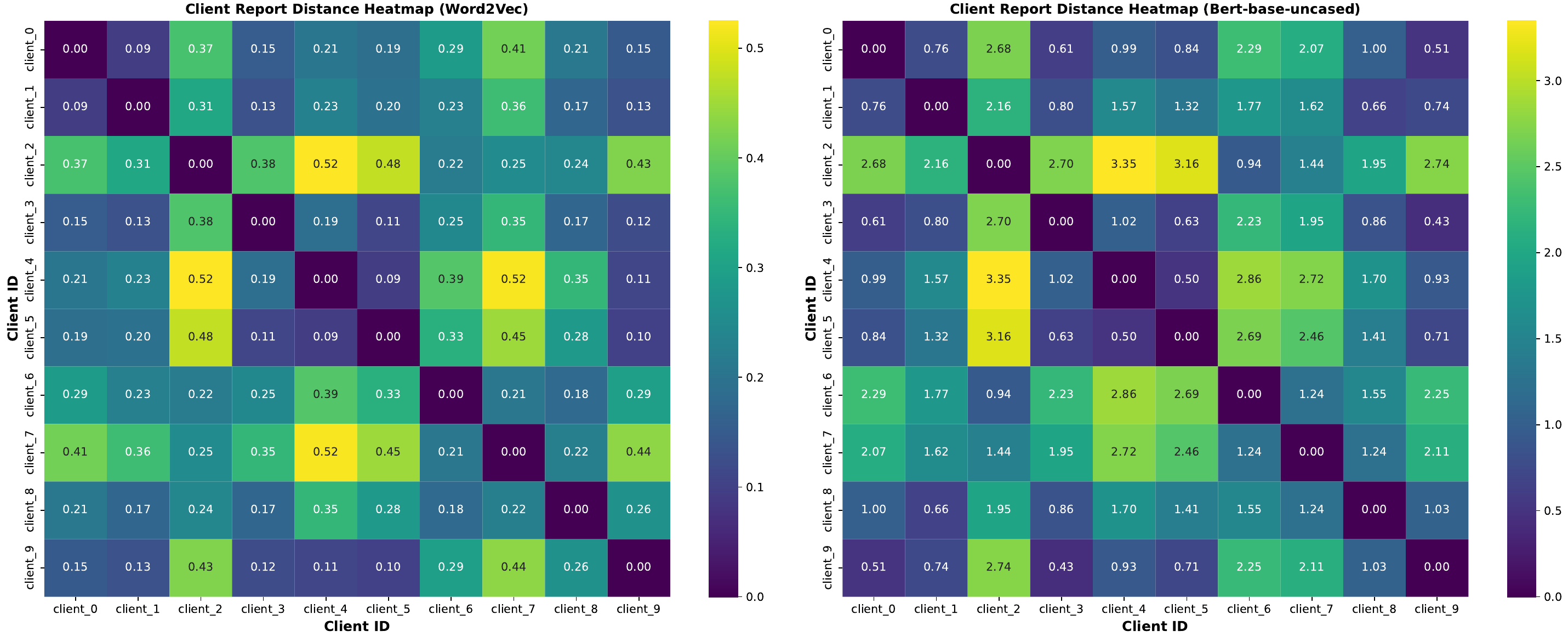}
    \vspace{-5mm}
    \caption{Client split by report-based clustering causes obvious data heterogeneity among clients at both word and semantics levels.}
    \label{fig:distance_matrices}
    \vspace{-3mm}
\end{figure}

\begin{table*}[t]
\begingroup

\centering
\caption{\small Comparisons with SOTA FL and MRG methods on the intra-domain test set.}
\footnotesize
\newcolumntype{C}{>{\centering\arraybackslash}X}%
\renewcommand\arraystretch{1.1}
\scalebox{0.86}{
\begin{tabular}{llcccccccc||cccccccc}
\toprule	 
\multirow{2}{*}{\textbf{Model}} & \multirow{2}{*}{\textbf{Comm.}} & \multicolumn{8}{c}{\textbf{FL-MRG (Cluster)}} & \multicolumn{8}{c}{\textbf{FL-MRG (Random)}} \\ 
\cmidrule(r){3-10} \cmidrule(r){11-18} 
& & BL1 & BL2 & BL3 & BL4 & ROU &F1 &REC &PRE & BL1 & BL2 & BL3 & BL4  & ROU &F1 &REC &PRE \\
\midrule
\multicolumn{18}{l}{\textit{\textbf{Federated \& MRG Baseline}}} \\
Centralized  & - & 32.70 & 20.40 & 13.90 & 13.43 & 27.71 & 29.9 & 27.8 & 37.8 & 32.70 & 20.40 & 13.90 & 13.43 & 27.71 & 29.9 & 27.8 & 37.8\\
Transformer\cite{chen2020generating} &63M & 27.05 & 17.00 & 11.43 & 8.16 & 26.54 & 16.1 & 13.8 & 23.3 & 27.68 & 17.48 & 11.86 & 8.54 & 26.82 & 19.0 & 16.0 & 28.3\\
R2Gen \cite{chen2020generating} &82M & 26.93 & 16.80 & 11.33 & 8.15 & 25.94 & 17.3 & 14.7 & 25.4 & 27.95 & 17.62 & 11.90 & 8.54 & 26.73 & 20.8 & 17.2 & 32.1\\
WCL \cite{yan2021weakly} &203M & 37.28 & 23.02 & 15.31 & 10.87 & 26.60 & 32.0 & \underline{29.6} & 41.0 & 32.54 & 18.63 & 12.24 & 8.99 & 26.13 & 27.8 & 31.3 & 37.5\\
DCL \cite{li2023dynamic} &203M & 36.96 & 22.59 & 15.10 & 10.86 & 26.75 & 28.4 & 26.2 & 36.3 & 37.11 & 23.46 & 15.42 & 11.22 & 27.08 & 29.6 & 27.1 & 38.7\\
\cmidrule(r){1-2} \cmidrule(r){3-7}  \cmidrule(r){8-10} \cmidrule(r){11-15} \cmidrule(r){16-18} 
\multicolumn{18}{l}{\textit{\textbf{LLM-Driven MRG}}} \\
R2GenGPT\cite{wang2023r2gengpt} &51M & 39.40 & 24.94 &\underline{17.01} & \underline{12.27} & \underline{27.22} & 26.6 & 23.8  & 35.7 & 38.78 & 24.62 & 16.87 & 12.19 & 27.04 & 29.1 & 27.3 & 36.9\\
PromptMRG \cite{jin2023promptmrg} &59M & 38.23 & 24.05 & 16.43 & 11.79 & 26.70 & \underline{30.7} & 26.4 & \underline{43.2} & 39.20 & 24.87 & 17.04 & 12.26 & 27.14 & \underline{32.2} & \underline{28.5} & \underline{43.3} \\
Fed-AdaLoRA \cite{zhang2023adlora} &57M & \underline{39.75} & \underline{24.90} & 16.85 & 12.03 & 27.10 & 29.9 & 27.2 & 38.7 & \underline{40.31} & 25.42 & 17.30 & 12.40 & 27.30 & 30.8 & 28.4 & 39.0 \\
FedPara \cite{hyeon2021fedpara} &76M & 38.48 & 24.16 & 16.38 & 11.74 & 26.13 & 26.4 & 23.8 & 35.0 & 39.78 & 25.03 & 16.99 & 12.19 & 27.11 & 27.3 & 23.9 & 37.5 \\
Fed-LoRA \cite{hu2021lora} &59M & 39.30 & 24.68 & 16.75 & 12.01 & 26.82 & 25.1 & 21.8 & 35.2 & 40.13 & \underline{25.50} & \underline{17.45} & \textbf{12.57} & \textbf{27.34} & 30.0 & 28.0 & 38.0 \\
Fed-Prefix \cite{li2021prefix} &103M & 38.46 & 24.10 & 16.43 & 11.90 & 26.77 & 26.9 & 24.1 & 35.5 & 39.57 & 24.88 & 16.94 & 12.19 & 27.06 & 28.0 & 25.4 & 36.3 \\
Fed-Prompt \cite{lester202prompt} &102M & 38.27 & 23.93 & 16.21 & 11.65 & 26.54 & 28.4 & 25.7 & 37.1 & 39.11 & 24.73 & 16.92 & 12.21 & 27.11 & 29.4 & 26.3 & 38.7 \\
Fed-Vera  \cite{kopiczko2023vera} &51M & 38.03 & 23.89 & 16.22 & 11.64 & 26.58 & 25.4  & 23.1 & 33.8 & 39.10 & 24.63 & 16.76 & 12.12 & 27.03 & 27.0 & 24.0 & 36.9 \\
FedDAT  \cite{chen2023feddat} &88M & 35.29  & 21.95 & 14.92  & 10.69  & 25.44  & 17.7 & 15.1  & 25.9 & 38.69  &  24.41 &  16.56 & 11.84  & 26.63  & 29.1  & 26.1  &  38.8  \\
\cmidrule(r){1-2} \cmidrule(r){3-7}  \cmidrule(r){8-10} \cmidrule(r){11-15} \cmidrule(r){16-18} 
\textbf{FedMRG (Ours)} &59M & \textbf{40.19} & \textbf{25.52} & \textbf{17.55} & \textbf{12.68} & \textbf{27.55} & \textbf{33.9} & \textbf{30.2} & \textbf{45.4} & \textbf{40.48} & \textbf{25.80} & \textbf{17.80} & \underline{12.36} & \underline{27.33} & \textbf{35.6} & \textbf{32.4} & \textbf{45.8} \\
\bottomrule
\end{tabular}
}
\vspace{-3mm}
\label{tab:results_sota}
\endgroup
\end{table*}

\begin{table*}[t]
\begingroup

\centering
\caption{\small Comparisons with SOTA FL and MRG methods on the unseen domain test.}
\footnotesize
\newcolumntype{C}{>{\centering\arraybackslash}X}%
\renewcommand\arraystretch{1.1}
\scalebox{0.86}{
\begin{tabular}{llcccccccc||cccccccc}
\toprule
\multirow{2}{*}{\textbf{Model}} & \multirow{2}{*}{\textbf{Comm.}} & \multicolumn{8}{c}{\textbf{FL-MRG (Cluster)}} & \multicolumn{8}{c}{\textbf{FL-MRG (Random)}} \\ 
\cmidrule(r){3-10} \cmidrule(r){11-18} 
& & BL1 & BL2 & BL3 & BL4 & ROU &F1 &REC &PRE & BL1 & BL2 & BL3 & BL4  & ROU &F1 &REC &PRE \\
\midrule
\multicolumn{18}{l}{\textit{\textbf{Federated \& MRG Baseline}}} \\
Centralized &-  & 39.56 & 22.39 & 13.42 & 8.55 & 28.85 & 15.2 & 15.2 & 15.8 & 39.56 & 22.39 & 13.42 & 8.55 & 28.85 & 15.2 & 15.2 & 15.8 \\
Transformer\cite{chen2020generating} &63M  & 32.04 & 18.62 & 11.25 & 6.66 & 26.07 & 13.62 & 13.44 & 14.20 & 34.15 & 19.10 & 12.06 & 7.73 & 26.10 & 14.2 & 14.1 & 14.6\\
R2Gen \cite{chen2020generating} &82M & 32.50 & 17.78 & 10.14 & 6.20 & 25.68 & 13.4 & 13.8 & 13.2 & 35.91 & 20.45 & 12.06 & 7.50 & 26.83 & 13.3 & 13.1 & 13.7\\
WCL \cite{yan2021weakly} &203M & 32.23 & 16.64 & 8.80 & 5.05 & 21.99 & 13.46 & 13.26 & 6.30 & 36.31 & 21.58 & 12.33 &8.01 & 27.53 & 13.5 & 13.5 & 14.0 \\
DCL \cite{li2023dynamic} &203M & 31.36 & 16.53 & 9.06 & 5.36 & 23.26 & 13.27 & 13.18 & 14.06 & 35.40 & 19.92 & 11.56 & 7.15 & 25.48 & 13.0 & 13.0 & 13.4 \\
\cmidrule(r){1-2} \cmidrule(r){3-10}  \cmidrule(r){11-18} 
\multicolumn{18}{l}{\textit{\textbf{LLM-Driven MRG}}} \\
R2GenGPT\cite{wang2023r2gengpt} &51M & 38.68 & 22.22 & 13.52 & 8.73 & 26.43 & 13.19 & 12.80 & 14.34 & {39.71} & {23.20} & 14.11 & 9.06 & 26.56 & 13.0 & 12.7 & 14.0 \\
PromptMRG \cite{jin2023promptmrg}  &59M & \underline{42.18} & \underline{25.33} & \underline{16.28} & \underline{11.05} & \textbf{30.61} & \underline{18.17} & \underline{15.76} & \underline{16.39} & \underline{42.05} & \underline{25.10} & \underline{15.99} & \underline{10.79} & \textbf{29.99} & \underline{17.0} & \underline{14.9} & \underline{15.4} \\
Fed-AdaLoRA \cite{zhang2023adlora} &57M & 36.93 & 21.08 & 12.71 & 8.11 & 25.66 & 13.01 & 12.78 & 13.91 & 38.40 & 22.04 & 13.17 & 8.29 & 26.28 & 12.9 & 12.7 & 13.8  \\
FedPara \cite{hyeon2021fedpara} &76M & 37.40 & 21.17 & 12.45 & 7.76 & 25.65 & 11.71 & 11.49 & 12.36 & 38.10 & 21.82 & 13.06 & 8.26 & 26.27 & 12.4 & 12.1 & 13.3  \\
Fed-LoRA \cite{hu2021lora} &59M  & 38.04 & 21.74 & 13.12 & 8.34 & 25.74 & 12.69 & 12.38 & 13.71 & 39.88 & 23.15 & 14.04 & 9.00 & 27.10 & 14.0 & 13.7 & 15.0 \\
Fed-Prefix \cite{li2021prefix} &103M & 37.68 & 21.63 & 13.19 & 8.53 & 25.82 & 13.07 & 12.76 & 14.08 & 37.85 & 21.62 & 12.98 & 8.24 & 26.17 & 12.5 & 12.3 & 13.4 \\
Fed-Prompt \cite{lester202prompt} &102M & 36.80 & 20.78 & 12.37 & 7.80 & 25.53 & 12.38 & 12.12 & 13.27 & 39.42 & 22.71 & 13.63 & 8.64 & 26.47 & 13.2 & 12.9 & 14.2 \\
Fed-Vera \cite{kopiczko2023vera} &51M  & 37.23 & 21.05 & 12.54 & 7.96 & 25.38 & 12.68 & 12.35 & 13.60  & 39.95 & 23.28 & 14.30 & 9.26 & 26.69 & 12.7 & 12.4 & 13.7 \\
FedDAT \cite{chen2023feddat} &88M  & 36.16  & 21.30  & 13.62 & 9.19 & 26.02 & 11.11 & 10.92 & 11.69  & 38.53 & 22.32 & 13.41 & 8.49 & 26.19 & 13.9 & 13.7 & 14.9 \\
\cmidrule(r){1-2} \cmidrule(r){3-10}  \cmidrule(r){11-18} 
\textbf{FedMRG(Ours)} &59M & \textbf{43.55} & \textbf{26.38} & \textbf{17.30} & \textbf{11.95} & \underline{29.73} & \textbf{19.69} & \textbf{17.22} & \textbf{17.86} & \textbf{42.99} & \textbf{25.70} & \textbf{16.45} & \textbf{11.15} & \underline{29.66} & \textbf{18.1} & \textbf{16.3} & \textbf{16.6} \\
\bottomrule
\end{tabular}
}
    \vspace{-3mm}
\label{tab:results_unseen}
\endgroup
\end{table*}

\subsection{Dataset Settings}
\subsubsection{Datasets and Federated Benchmark Design} 
We develop a publicly-accessible federated MRG benchmark (FL-MRG) to simulate realistic federated learning environments for medical report generation.
Our benchmark incorporates two established chest X-ray datasets: 1) MIMIC-CXR \cite{johnson2019mimic}, the largest available MRG dataset containing 276,778 images with corresponding reports after standardized preprocessing \cite{chen2020generating}; and 
2) IU X-Ray \cite{demner2016iucxr}, a widely adopted MRG evaluation dataset comprising 4,168 images following preprocessing procedures outlined in \cite{jin2023promptmrg}.} 
Specifically, we construct our federated MRG benchmark by partitioning the official MIMIC-CXR training set into ten distinct subsets using two approaches: patient-level random sampling (FL-MRG Random) and patient-level report-based clustering (FL-MRG Clustering).
Within FL-MRG, 90\% of each local dataset is allocated for training purposes, while the remaining 10\% serves as client-specific validation data.
To comprehensively evaluate FedMRG's effectiveness and generalization capabilities in federated learning scenarios, we implement two distinct evaluation protocols: in-domain testing using the original MIMIC-CXR test set and unseen domain testing using the complete IU X-Ray dataset.
To the best of our knowledge, FL-MRG represents the first benchmark specifically designed to simulate real-world federated chest X-ray environments, featuring a comprehensive evaluation across both intra-domain and inter-domain settings.

\subsubsection{Client Heterogeneity Discussion}
Through patient-level random sampling and report-based clustering approaches, we simulate realistic data heterogeneity and distribution patterns characteristic of real-world federated learning environments. The patient-level partitioning strategy reflects the natural distribution of medical data across diverse healthcare institutions, further enhanced by applying client-specific visual transformations — including blurring, contrast modifications, and brightness adjustments — to simulate imaging device variability across facilities, following methodologies established in \cite{shen2020modeling}.
FL-MRG (Random), implemented through stratified random sampling, ensures heterogeneous case distribution across subsets, reflecting the inherent unpredictability and diversity of distributed medical data sources while introducing moderate data heterogeneity.
Conversely, FL-MRG (Clustering), employing report-based clustering algorithms, strategically groups similar clinical cases, simulating the geographical concentration of specific conditions or diseases within particular regions or demographic populations. This approach generates more pronounced data heterogeneity across clients, as quantitatively demonstrated in Fig. \ref{fig:distance_matrices}.

\begin{table*}[t]
\begingroup

\centering
\caption{\small Evaluation of FedMRG and baseline methods under real-world federated settings. Results on internal test set and external test set (IU X-ray) are shown, respectively.}
\footnotesize
\newcolumntype{C}{>{\centering\arraybackslash}X}%
\renewcommand\arraystretch{1.1}
\scalebox{0.86}{
\begin{tabular}{llcccccccc||cccccccc}
\toprule	 
\multirow{2}{*}{\textbf{Model}} & \multirow{2}{*}{\textbf{Comm.}} & \multicolumn{8}{c}{\textbf{Internal Test (MIMIC-CXR \& Chexpert+)}} & \multicolumn{8}{c}{\textbf{External Test (IU X-ray)}} \\ 
\cmidrule(r){3-10} \cmidrule(r){11-18} 
& & BL1 & BL2 & BL3 & BL4 & ROU &F1 &REC &PRE & BL1 & BL2 & BL3 & BL4  & ROU &F1 &REC &PRE \\
\midrule
\multicolumn{18}{l}{\textit{\textbf{Federated \& MRG Baseline}}} \\
Centralized  & - & 21.39 & 13.45 & 9.20 & 6.49  & \textbf{24.92}  & \underline{21.76}  & 18.72  & 27.46  & 33.93 & 19.71  & 12.41  & 8.11 & 28.46 & \underline{13.23} & \underline{13.09} & \underline{13.58} \\
Transformer\cite{chen2020generating} &63M & 18.33 & 11.73 & 8.01 & 5.69  & \underline{24.82}  &  14.28  & 12.45 & 18.55 & 36.98 & 23.34 & 15.98  &  11.49  & 30.56 & 12.49  & 12.44 & 12.59 \\
R2Gen \cite{chen2020generating} &82M & 19.51  & 12.36  & 8.38  & 5.87  & \underline{24.82}  & 19.35  &  16.48  &  27.17 & 32.47  & 19.84 & 12.89 & 8.68  & 29.41 & 13.04 & 12.94 & 13.32 \\
WCL \cite{yan2021weakly} &203M & 28.68 & 16.77 & 10.21 & 6.51 & 22.63  & 15.09  & 13.50 & 20.69  & 26.94 & 14.99 & 8.48 & 5.25 & 21.85  & 5.44 & 5.52 & 5.52 \\
DCL \cite{li2023dynamic} &203M & 29.97  & 17.64  & 10.98  & 7.19  &  22.99 &  10.59 &  9.86 & 13.05 & 38.90 & 22.77 & 13.62 & 8.63  & 25.99  & 9.85  & 9.84  & 9.91 \\
\cmidrule(r){1-2} \cmidrule(r){3-7}  \cmidrule(r){8-10} \cmidrule(r){11-15} \cmidrule(r){16-18} 
\multicolumn{18}{l}{\textit{\textbf{LLM-Driven MRG}}} \\
R2GenGPT\cite{wang2023r2gengpt} &51M & 30.16 & 17.46 & 11.16 & 7.48 & 23.18 & 13.85 & 12.14 & 18.31 & 37.69 & 24.51 & 17.56 & 13.07  & 33.18  & 9.96 & 9.93  &  10.05 \\
PromptMRG \cite{jin2023promptmrg} &59M & \underline{34.57} & 20.68 & 13.29 & \underline{8.90}  & 24.08  &  12.65 & 10.73 &  17.99 & 37.01 & 23.01 & 15.65 & 11.18 & 28.16 & 10.03  & 9.97 & 10.20 \\
Fed-AdaLoRA \cite{zhang2023adlora} &57M & 31.72 & 18.55 & 11.89 & 7.98 & 23.32  & 15.87 & 13.74 & 21.78 & 37.61 & 24.51 & 17.66 & 13.18 & 33.12 & 10.07  &  10.01 & 10.24 \\
FedPara \cite{hyeon2021fedpara} &76M & 31.10 & 18.37  &  11.85 &  8.02 &  23.66 & 16.32  & 14.01 & 22.82 & 37.54  & 24.16  & 17.16  & 12.60  & 32.56  & 10.04  & 9.97 & 10.23 \\
Fed-LoRA \cite{hu2021lora} &59M & \textbf{35.29}  & 20.38  & 12.79  &  8.48 & 23.54 & 19.96  & 17.06  &  28.14 & 36.75 & 23.60 & 16.50  & 11.98  & 30.22  & 10.50  & 10.93  & 12.35 \\
Fed-Prefix \cite{li2021prefix} &103M & 32.92 & 19.55 & 12.52 & 8.45  & 23.72  & 19.08  &  16.39 &  26.43 & 39.61 & 25.64 & 18.06 & 13.23 & 32.16  & 10.42  & 10.28 & 10.77 \\
Fed-Prompt \cite{lester202prompt} &102M & 34.10 & \underline{20.79} & \underline{13.56} & \textbf{9.30}  & 24.75 & 17.71 & 15.46  & 24.10  & \textbf{43.80} & \underline{27.95}  & \underline{19.53}  & 14.19  & 29.81  & 10.21  & 10.11  & 10.50 \\
Fed-Vera  \cite{kopiczko2023vera} &51M & 32.60  &  19.18 & 12.43  & 8.42 & 23.56  & 14.88  &  13.23 & 19.39 & 39.01 & 25.90 & 18.90 & \underline{14.27} & \textbf{33.73} & 9.96 & 9.93 & 10.03 \\
FedDAT \cite{chen2023feddat} &88M & 33.01 & 19.27 & 12.26  & 8.09   & 23.58  & 21.17  & \underline{19.01}  & \underline{28.18} & 39.11  & 24.86 & 17.54  & 12.88 & 32.43 & 10.58 & 10.36 & 11.18 \\
\cmidrule(r){1-2} \cmidrule(r){3-7}  \cmidrule(r){8-10} \cmidrule(r){11-15} \cmidrule(r){16-18} 
\textbf{FedMRG (Ours)} &59M & \textbf{35.29}  & \textbf{21.14}  & \textbf{13.98} & \underline{8.90}  & 24.34  & \textbf{22.61}  &  \textbf{20.05}  & \textbf{29.76}  & \underline{42.49}  & \textbf{28.62}  & \textbf{20.55}  & \textbf{14.92}  & \underline{33.57}  & \textbf{15.37}  & \textbf{15.42}  & \textbf{16.18} \\
\bottomrule
\end{tabular}
}
\label{tab:results_realworld}
\endgroup
\vspace{-3mm}
\end{table*}

\subsubsection{Real-world Federated Setting}
To rigorously evaluate FedMRG's generalization capabilities and personalization efficacy under more realistic federated learning settings, we incorporate a comprehensive multi-source dataset by integrating CheXpert+ \cite{irvin2019chexpert}, comprising 224,316 chest radiographs (both frontal and lateral views) from 65,240 patients. Following meticulous quality assessment and clinical relevance filtering — excluding reports with fewer than 20 words and those lacking a structured ``findings" section — we extract approximately 55,000 high-quality image-report pairs from CheXpert+ and an equivalent volume from MIMIC-CXR \cite{johnson2019mimic}. We strategically partitioned each institutional dataset across three distinct federated clients, creating a genuine multi-institutional collaborative dataset that preserves the naturally occurring data heterogeneity intrinsic to different healthcare facilities, rather than relying solely on artificial transformations. This enhanced evaluation approach enables us to assess FedMRG's performance under conditions that closely mirror real-world clinical deployment scenarios across diverse medical institutions.

\subsubsection{Evaluation Metric} 
We evaluated model performance with both natural language generation (NLG) metrics and clinical efficacy (CE) metrics, following \cite{jin2023promptmrg}. 
The NLG metrics include BLEU (BL1-BL4) \cite{papineni2002bleu}, CIDEr (CID) \cite{vedantam2015cider}, and ROUGE-L (ROU) \cite{vedantam2015cider}. 
Additionally, following \cite{nicolson2023ce}, the CE metrics include example-based precision score (PRE), example-based recall score (REC), and example-based F1 score (F1), which are evaluated by converting reports into 14 disease classification labels using CheXbert \cite{smit2020combining}.

\subsection{Implementation Details}
We differentiate between two categories of language models based on parameter scale: Conventional Language Models (CLMs) with parameters under 1B, and Large Language Models (LLMs) exceeding this threshold.
This distinction is necessitated by the communication efficiency requirements inherent to federated learning settings \cite{rieke2020future}.
For CLMs, we designate all parameters as trainable and allow their participation in cross-client communications.
Conversely, for LLMs, the computational and bandwidth costs associated with transmitting and distributing their complete parameter sets are prohibitively expensive and practically infeasible, as substantiated by previous research \cite{chen2023feddat}.
Therefore, when utilizing LLMs, we maintain their parameters in a frozen state and exclude them from communication except during the initial model distribution phase.
Within our framework, the visual encoder and global adapters remain trainable and participate in communication, while personalized adapters, though trainable, are restricted to local updates without cross-client transmission.
Our implementation employs an ImageNet pre-trained ResNet-101 \cite{he2016deep} as the visual encoder and Llama2-7B-Chat-HF \cite{touvron2023llama} as the language decoder.
We optimize the model using AdamW \cite{loshchilov2019decoupled} with a weight decay coefficient of 0.05.
The learning rate is initialized at 5e-5 and modulated according to a cosine learning rate schedule.
The federated training process encompasses 1000 communication rounds, with each local model processing approximately 2\% of its client-specific dataset per round. We configure the training with a batch size of 16, standardize input images to $224\times224$ dimensions, and set the temperature parameter $\tau$ to 0.07.
The entire system was implemented using the PyTorch framework and trained on a single H800 GPU.

\subsection{Main Results}
\subsubsection{Comparison Method}
We conduct a comparative analysis of our FedMRG method with several state-of-the-art (SOTA) methods in the MRG field, namely Transformer \cite{chen2020generating}, R2Gen \cite{chen2020generating}, WCL \cite{yan2021weakly}, R2GenGPT \cite{wang2023r2gengpt}, DCL \cite{li2023dynamic}, and PromptMRG \cite{jin2023promptmrg}. 
Notably, the R2GenGPT model employs a frozen LLM, Llama2-7B, as its decoder component. 
Additionally, our experiment encompasses several federated and LLM efficient adaptation algorithms, including the classic FedAvg \cite{mcmahan2017fedavg}, and AdaLora \cite{zhang2023adlora}, FedPara \cite{hyeon2021fedpara}, LoRA\cite{hu2021lora}, Prefix\cite{li2021prefix}, Prompt\cite{lester202prompt}, Vera\cite{kopiczko2023vera}, and FedDAT \cite{chen2023feddat}.
During our experiments, we treated FedAvg as the default FL algorithm for MRG methods and the Transformer as the baseline method.
The top two scores are denoted in \textbf{bold} and by \underline{underlining}.

\subsubsection{Inside-federation Comparison}
As shown in Table \ref{tab:results_sota}, we first set the Transformer model as the baseline to investigate the performance gap between centralized and federated learning scenarios. 
Notably, the Transformer's performance decreases substantially within the federated setting compared to centralized training, highlighting the significant challenges that MRG models face in a data-isolated federated context.
When combining FedAvg with state-of-the-art (SOTA) MRG methods, we observe significant performance improvements over the baseline model, demonstrating their effective designs for MRG tasks. 
Both WCL and DCL introduce representation learning, enhancing the performance of basic MRG models. 
Additionally, our results show that LLM-driven MRG models consistently outperform conventional MRG methods, emphasizing the importance of integrating LLMs into MRG, especially in federated scenarios.
As expected, FedMRG stands out due to its design of representation learning and comprehensive adapter-based personalization approach, consistently outperforming established methods across both inside-client and MIMIC tests.
These results validate the effectiveness of FedMRG in handling the inherent multi-modal data heterogeneity in federated settings while maintaining low communication costs. 
This makes it particularly suitable for training LLM-driven MRG models in federated healthcare environments.

\subsubsection{Outside-federation Comparison}
Robust generalization capability is essential for effective FedMRG methods. To evaluate this aspect, we conducted a challenging unseen domain generalization test using the complete IU X-Ray dataset, following the protocol established in \cite{jin2023promptmrg}.
As shown in Table \ref{tab:results_unseen}, FedMRG maintains its superior performance compared to other SOTA methods, demonstrating its resilience to domain shifts between the MIMIC and IU X-Ray datasets.
In this outside-federation comparison, we observe a general performance decline relative to inside-federation results, highlighting the significant impact of data heterogeneity.
Notably, the LLM-driven MRG models powered by Llama2 show a smaller performance drop compared to conventional methods. This indicates the strategic advantage of incorporating advanced LLMs within MRG frameworks, an approach that has proven particularly effective for FedMRG. 
Meanwhile, PromptMRG performs exceptionally well in cross-domain settings, likely due to its specialized prompt design that captures diagnostic information. 
Our approach extends this concept by integrating representation learning to capture discriminative knowledge, thereby achieving superior performance across both CE and NLG metrics. In summary, these results demonstrate FedMRG's ability to deliver both personalized and generalizable MRG models while confirming the effectiveness of its design in addressing the complex challenges inherent to federated medical report generation.

\subsubsection{Clinical Efficiency Investigation}
We have conducted a comprehensive evaluation using CE metrics, which are essential for assessing the practical utility of MRG models in real-world healthcare environments.
As demonstrated in Table \ref{tab:results_sota} and Table \ref{tab:results_unseen}, our method consistently outperforms other models in CE metrics across both inside-federation and outside-federation tests. 
This superior clinical performance can be attributed primarily to our integration of representation learning modules and the explicitly designed diagnosis prompts. By providing the LLM decoder with more discriminative visual features and structured diagnosis information as prompts, our approach enables enhanced diagnostic awareness within the generation process. This improved clinical context allows the model to generate more accurate and clinically relevant reports, ultimately enhancing diagnostic efficiency in practical medical settings.

\subsubsection{Real-world Federated Setting}
Table \ref{tab:results_realworld} presents our evaluation of FedMRG under more challenging real-world federated conditions, where we integrate both MIMIC-CXR and CheXpert+ datasets across multiple clients to simulate genuine cross-institutional heterogeneity. The results reveal several notable patterns. First, FedMRG consistently outperforms all baseline and state-of-the-art methods on both internal validation (MIMIC-CXR \& CheXpert+) and external testing (IU X-Ray), demonstrating robust generalization across diverse data sources. For internal tests, our approach achieves the highest scores in both language generation metrics and clinical efficacy measures, with particularly significant improvements in precision and recall compared to the second-best method.

On the unseen IU X-Ray dataset, FedMRG also achieves comparable or superior performance compared to baseline methods. These results validate that FedMRG not only addresses the technical challenges of federated medical report generation but also offers a practical solution for real-world clinical deployment across diverse healthcare institutions.

\begin{table}[t]
\begingroup

\centering
\caption{\small Ablation study on the MIMIC-CXR test.}
\footnotesize
\newcolumntype{C}{>{\centering\arraybackslash}X}%
\renewcommand\arraystretch{1}
\scalebox{0.9}{
\begin{tabular}{lcccccccccc}
\toprule	 
\multirow{1}{*}{\textbf{Model}} & BL1 & BL2 & BL3 & BL4 & ROU  &F1 &REC &PRE \\
\midrule
\textbf{FedMRG} & \textbf{40.2} & \textbf{25.5} & \textbf{17.6} & \textbf{12.7} & \textbf{27.6} & \textbf{33.9} & \textbf{30.2} & \textbf{45.4}  \\
\cmidrule(r){1-1} \cmidrule(r){2-6} \cmidrule(r){7-9}
\textbf{$\cdot$ w/o $\mathcal{A}_s$} & 38.7 & 24.2 & 16.4 & 11.6 & 26.7 & 32.0 & \underline{29.1} & 41.8 \\
\textbf{$\cdot$ w/o $\mathcal{L}_{hcl}$} & 37.9 & 23.7 & 16.1 & 11.5 & 26.7 & 30.8 & 27.5 & 41.3 \\
\textbf{$\cdot$ w/o $p$} & 38.2 & 24.1 & 16.4 & 11.8 & 26.8 & 29.3 & 26.3 & 38.5 \\
\textbf{$\cdot$ w/o $\mathcal{L}_{g2l}$} & \underline{39.6} & \underline{24.9} & \underline{17.0} & \underline{12.2} & \underline{27.2} & \underline{31.5} & 27.2 & \underline{44.2} \\
\textbf{$\cdot$ w/o $\mathcal{L}_{l2g}$} & \underline{39.6} & \underline{24.9} & 16.9 & \underline{12.2} & 27.1 & 30.8 & 26.3 & 43.9 \\
\bottomrule
\end{tabular}
}
\label{tab:ablation}
\endgroup
\vspace{-3mm}
\end{table}

\subsection{Ablation Studies}
\subsubsection{Effects of Each Component}
We have conducted an ablation study to analyze the contributions of individual components within the FedMRG framework, with results presented in Table \ref{tab:ablation}. 
This systematic evaluation involved removing specific elements sequentially and measuring the resultant impact on performance metrics.
As evidenced in the results, removing the representation loss $\mathcal{L}_{hcl}$ leads to a significant performance decline across all metrics, highlighting its essential role in enhancing the quality and clinical relevance of generated reports.
Similarly, excluding the specialized adapter $A_s$ and prompts $p$ results in diminished performance, particularly in NLG metrics, demonstrating their importance in aligning linguistic patterns between global and local knowledge bases.
Additionally, the individual contributions of knowledge transfer guidance functions $\mathcal{L}_{g2l}$ and $\mathcal{L}_{l2g}$ prove to be substantial. 
Their removal causes consistent performance reductions across metrics, illustrating their synergistic effect in optimizing model performance and confirming the effectiveness of our mutual knowledge boosting mechanism. These experimental findings confirm that each component of FedMRG contributes meaningfully to the model's overall effectiveness. The observed performance degradation when components are removed demonstrates the integrated nature of FedMRG's architecture, where each element serves a specific purpose in enabling the generation of accurate and clinically valuable medical reports.

\begin{figure}[t]
    \centering
    \includegraphics[width=1\linewidth]{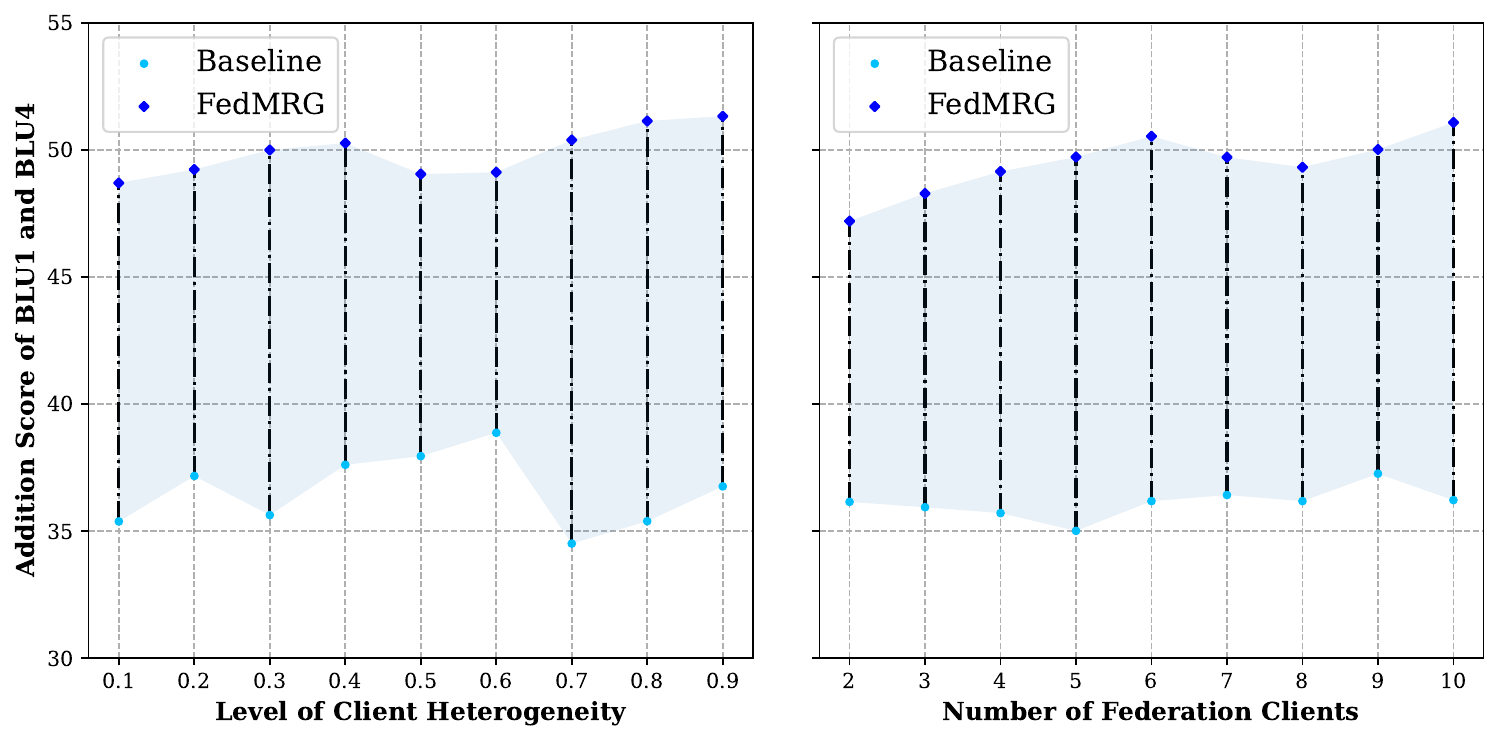}
    \vspace{-5mm}
    \caption{Performance of the baseline and FedMRG towards different levels of client heterogeneity and different numbers of clients.}
    \vspace{-2mm}
    \label{fig:node}
\end{figure}

\begin{figure}[t]
    \centering
    \includegraphics[width=1\linewidth]{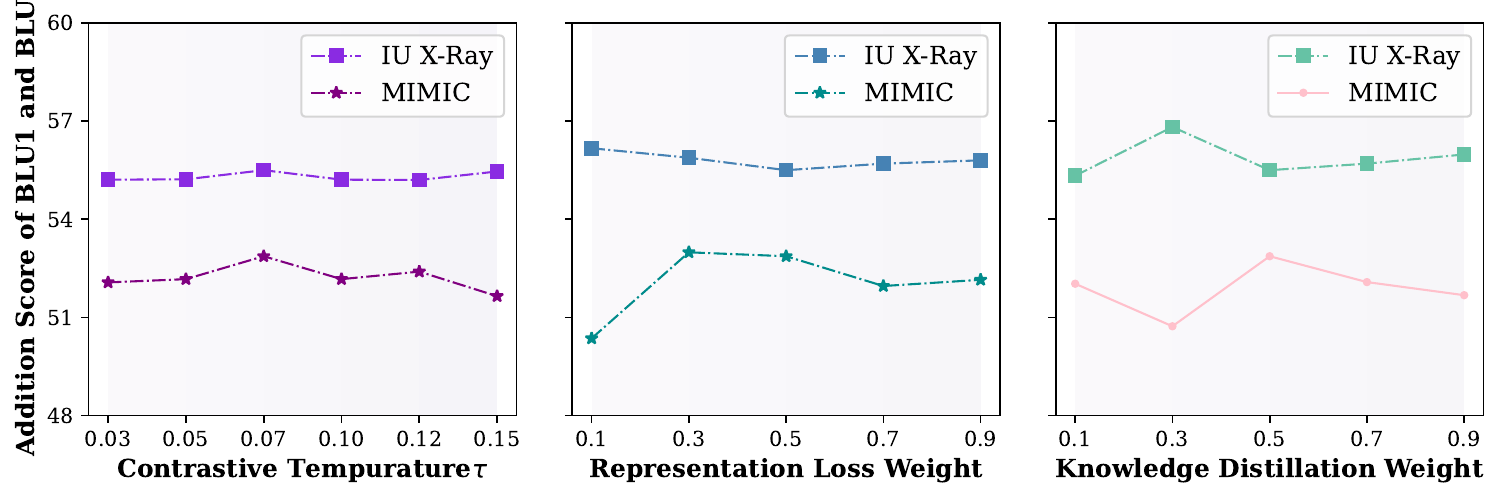}
    \vspace{-5mm}
    \caption{Effects on hyper-parameters on the model performance.}
    \vspace{-3mm}
    \label{fig:hama}
\end{figure}

\begin{figure*}[t]
    \centering
    \includegraphics[width=1\textwidth]{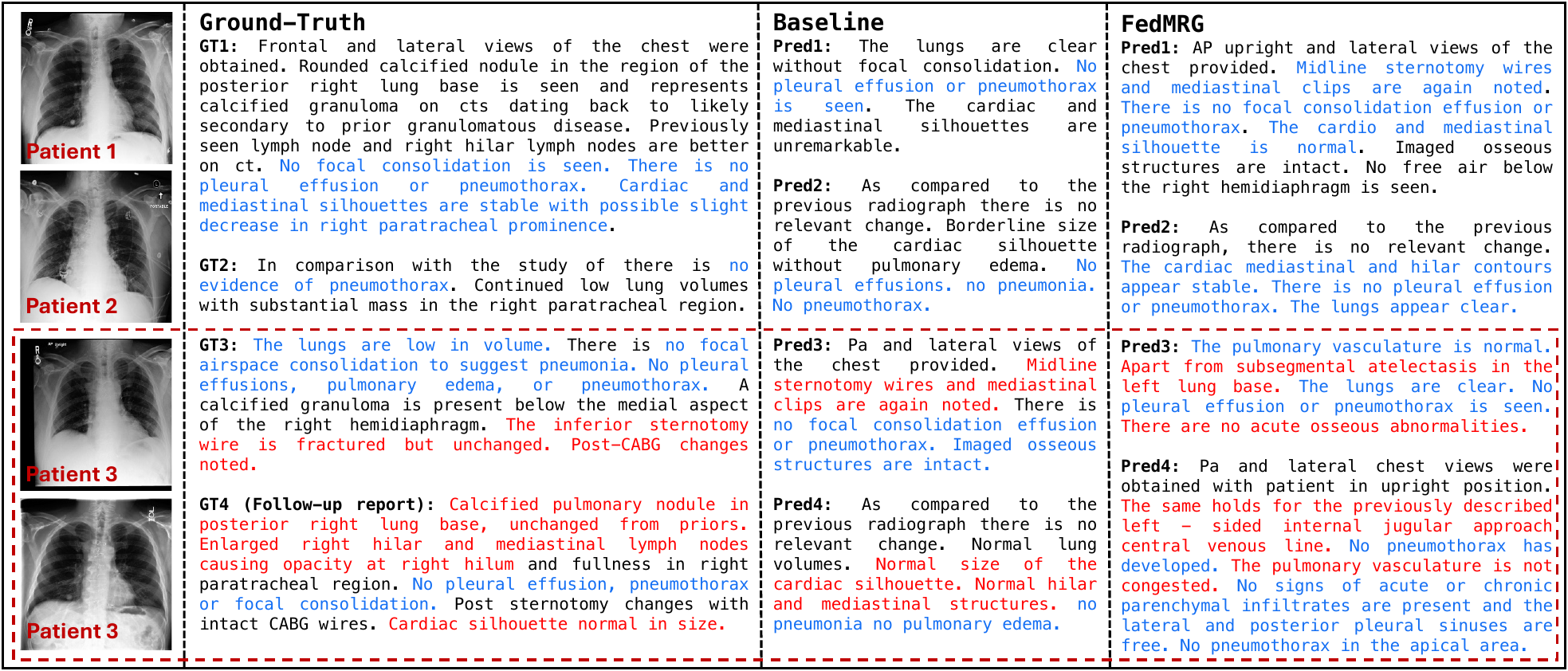}
    \vspace{-5mm}
    \caption{Qualitative examples of the baseline and FedMRG. \textcolor{blue}{Blue font} indicates consistent content with the ground-truth, while \textcolor{red}{red font} indicates patient-aware content.}
    \vspace{-3mm}
    \label{fig:generation}
\end{figure*}

\subsubsection{Investigation on Class Heterogeneity and Federation Scale}
We have further evaluated our model's performance through intra-domain tests across various federated configurations, examining the effects of client heterogeneity levels and federation size, as illustrated in Fig. \ref{fig:node}.
First, we distribute data across ten clients using the Dirichlet method, with concentration parameter $\alpha$ ranging from 0.1 to 0.9, while maintaining client data samples between 4000 and 7000 to avoid empty-sample scenarios.
Subsequently, we explore the impact of federation scale by varying the number of participating clients from 2 to 10, aiming to understand how participant count and data distribution influence model performance. Our analysis revealed that the baseline model exhibited significant performance fluctuations across different heterogeneity levels and federation sizes.
Specifically, performance scores ranged from 34.51 to 38.87 (w/ std of 1.358) and 35.01 to 37.26 (w/ std of 0.561), respectively. 
These variations indicate sensitivity to federation configuration and highlight the baseline model's limitations in scaling capabilities and adaptation to client heterogeneity.
In contrast, FedMRG demonstrated more consistent performance with notable improvements as federation scale increased and heterogeneity decreased.
Results ranged from 48.69 to 51.33 (w/ std of 0.897) and 47.2 to 51.08 (w/ std of 1.096), respectively. 
This performance stability and progressive enhancement underscore FedMRG's superior scalability and effectiveness in leveraging expanded federations while accommodating client data heterogeneity.

\subsubsection{Impacts of Hyper-parameters on Model Performance}
As illustrated in Fig. \ref{fig:hama}, we identify clear performance trends related to temperature parameter $\tau$, representation loss weight, and knowledge distillation weight.
The temperature parameter $\tau$ functions as a scaling factor that controls the separation between positive and negative examples in contrastive learning. Our experiments demonstrate an inverse relationship between temperature values and model performance. 
Higher temperature settings tend to smooth the feature distribution, resulting in less discriminative representations. Consequently, increasing temperature values correlate with performance degradation across both IU X-Ray and MIMIC datasets. For representation loss weight, we find that careful calibration is essential for optimal performance. 
As shown in the middle panel of Fig. \ref{fig:hama}, intermediate values of representation loss weight produce superior results. 
Both excessively low and high weights compromise the model's representation learning capabilities, indicating that moderate emphasis on representation loss achieves an effective balance between task-specific learning and feature representation quality.
Regarding knowledge distillation weight, the right panel of Fig. \ref{fig:hama} reveals an optimal range for this parameter. 
While knowledge distillation facilitates information transfer between model components, either too low or too high a weight can adversely affect performance. 
Our analysis confirms that moderate knowledge distillation weights achieve the best results, ensuring effective knowledge transfer while preserving the integrity of primary learning objectives.
These findings emphasize the importance of careful hyper-parameter tuning to maintain an appropriate balance between competing learning objectives. 
Optimal configuration of temperature $\tau$, representation loss weight, and knowledge distillation weight proves critical for maximizing model performance in federated MRG.

\subsubsection{Comparison in Generated Reports}
As shown in Fig.~\ref{fig:generation}, we provide qualitative examples from the baseline model and FedMRG across multiple patients. In the comparison of different case reports, the baseline model tends to generate more generalized predictions, lacking specific descriptive elements. For Patient 1 and 2, our model demonstrates better alignment with ground truth findings, correctly capturing details like ``midline sternotomy wires'' and the stability of cardiac contours. For Patient 3, the baseline struggles with consistency, missing the sternotomy wire in some reports while mentioning it in others. In contrast, FedMRG shows stronger continuity across multiple scans, noting patient-specific findings like ``subsegmental atelectasis'' and maintaining awareness of the patient's history through references such as ``previously described left-sided internal jugular approach central venous line."

Our model's enhanced performance can be attributed to the specific design of representation learning that captures potential correlations among different views, along with our dual-adapter mechanism that balances global knowledge with patient-specific features. The qualitative results suggest that FedMRG offers improved clinical detail, report consistency, and patient-specific context awareness, which may translate to better patient-friendly clinical utility in real-world settings.

%% file: 5_conclusion.tex
\section{Conclusion and Discussion}
\label{sec:conclu}

In this paper, we propose FedMRG, an effective approach to transform LLMs for communication-efficient MRG, representing a significant stride towards addressing the complex challenges of data privacy and accessibility in medical report generation under federated learning settings. FedMRG employed hierarchical contrastive learning and prompting to learn globally generalizable feature representations while preserving instance-level differences, and leveraged hybrid expert fusion through dual adapters to achieve effective communication-efficient LLM adaptation. Our extensive experiments highlight FedMRG's proficiency in leveraging diverse, multi-center data to produce accurate and clinically relevant reports.

While FedMRG demonstrates strong performance, we acknowledge several limitations that warrant future investigation. Our approach simulates heterogeneity through visual transformations and report-based clustering, but real-world medical data heterogeneity is more complex, encompassing variations in disease prevalence, patient demographics, and institutional reporting preferences. Additionally, our framework assumes synchronized client participation, which may not always be feasible in clinical environments with varying operational schedules.

FedMRG balances the competing demands of data privacy, communication efficiency, and report quality while demonstrating robust generalization capabilities across both seen and unseen domains. The performance on the IU X-Ray dataset highlights the model's resilience to domain shift, which is crucial for clinical deployment where models must perform reliably across diverse settings. Furthermore, the communication efficiency achieved through low-rank adaptation makes our approach particularly suitable for scaling to larger federations of medical institutions, with our experiments suggesting that performance generally improves as more clients join the federation. By effectively addressing the unique challenges of multi-modal data heterogeneity in federated settings, FedMRG paves the way for future research in secure, collaborative medical data utilization.